\documentclass[submission]{eptcs}
\usepackage{underscore}           % Only needed if you use pdflatex.
\usepackage{cite}
\usepackage[all]{foreign}
\usepackage{amsthm}
\usepackage[inline]{enumitem}
\usepackage{hhline}
\usepackage{float}
\usepackage{array}
\usepackage{subcaption}
\usepackage{pdfpages}

\usepackage{etoolbox}
\newcommand*{\defemph}[1]{\ensuremath{\mathsf{#1}}}

\newcommand*{\agentstyle}[1]{{\ensuremath{\uppercase{\defemph{#1}}}}}
\newcommand*{\agentstyleMin}[1]{{\ensuremath{\lowercase{\defemph{#1}}}}}%}
\newcommand{\agent}[1]{%
  \ifstrequal{#1}{i}%
             {\ensuremath{\lowercase{\defemph{#1}}}}%
             {\ifstrequal{#1}{A}{\agentstyle{#1}}{%
\ifstrequal{#1}{a}{\agentstyle{#1}}{%
\ifstrequal{#1}{B}{\agentstyle{#1}}{%
\ifstrequal{#1}{b}{\agentstyle{#1}}{%
\ifstrequal{#1}{C}{\agentstyle{#1}}{%
\ifstrequal{#1}{c}{\agentstyle{#1}}{%
\ifstrequal{#1}{D}{\agentstyle{#1}}{%
\ifstrequal{#1}{d}{\agentstyle{#1}}{%
\ifstrequal{#1}{E}{\agentstyle{#1}}{%
\ifstrequal{#1}{e}{\agentstyle{#1}}{%
\ifstrequal{#1}{ag}{\agentstyleMin{#1}}{%
\ifstrequal{#1}{x}{\agentstyleMin{#1}}{%
\ifstrequal{#1}{y}{\agentstyleMin{#1}}{%
\ifstrequal{#1}{AG}{\agentstyleMin{#1}}{%
\ifstrequal{#1}{ag_1}{\agentstyleMin{#1}}{%
\ifstrequal{#1}{ag_2}{\agentstyleMin{#1}}{%
\ifstrequal{#1}{ag_i}{\agentstyleMin{#1}}{%
\ifstrequal{#1}{ag_n}{\agentstyleMin{#1}}{??
}}}}}}}}}}}}}}}}}}}%
}

\newcommand*{\C}{\textbf{C}}

\newcommand*{\cAlpha}[1]{\ensuremath{\mathbf{C}_\alpha{#1}}}
\newcommand*{\eAlpha}[1]{\ensuremath{\mathbf{E}_\alpha{#1}}}
\newcommand*{\eAlphaIter}[2]{\ensuremath{\mathbf{E}^{#1}_\alpha{#2}}}
\newcommand*{\bB}[2]{\mathbf{B}_{\agent{#1}}{#2}}
\renewcommand*{\b}[1]{\ensuremath{\mathbf{B_{\agent{#1}}}}}
\newcommand*{\del}{DEL}
\newcommand*{\brel}[1]{\ensuremath{\calB_{\defemph{#1}}}}

\newcommand*{\distract}[2]{%
\ifstrequal{#2}{}%
{\ensuremath{\mathtt{distract}(\agent{#1})}}%
{\ensuremath{\mathtt{distract}(\agent{#1})\langle\agent{#2}\rangle}}%
}
\newcommand*{\open}[1]{%
\ifstrequal{#1}{}%
{\ensuremath{\mathtt{open}}}%
{\ensuremath{\mathtt{open}\tuple{\agent{#1}}}}%
}
\newcommand*{\shout}[1]{%
\ifstrequal{#1}{}%
{\ensuremath{\mathtt{shout\_tails}}}%
{\ensuremath{\mathtt{shout\_tails}\tuple{\agent{#1}}}}%
}
\newcommand*{\signal}[2]{%
\ifstrequal{#2}{}%
{\ensuremath{\mathtt{signal}(\agent{#1})}}%
{\ensuremath{\mathtt{signal}(\agent{#1})\langle\agent{#2}\rangle}}%
}
\newcommand*{\peek}[1]{%
\ifstrequal{#1}{}%
{\ensuremath{\mathtt{peek}}}%
{\ensuremath{\mathtt{peek}\tuple{\agent{#1}}}}%
}
\newcommand*{\tell}[2]{%
\ifstrequal{#2}{}%
{\ensuremath{\mathtt{tell}(\agent{#1})}}%
{\ensuremath{\mathtt{tell}(\agent{#1})\langle\agent{#2}\rangle}}%
}
\newcommand*{\flip}[1]{%
\ifstrequal{#1}{}%
{\ensuremath{\mathtt{flip}}}%
{\ensuremath{\mathtt{flip}\tuple{\agent{#1}}}}%
}

\newcommand*{\ourL}{\ensuremath{m\mathcal{A}^\rho}}
\newcommand*{\mAL}{\ensuremath{m\mathcal{A}^*}}

\newcommand*{\ck}{common knowledge}

\newcommand*{\wf}{well-founded}

\newcommand*{\nwf}{non-\wf}
\newcommand*{\posS}{possibilities}

\newcommand*{\lAG}{\ensuremath{\mathcal{L}_{\sAG}}}
\newcommand*{\lag}{\lAG}
\newcommand*{\lagC}{\ensuremath{\lAG^{\C}}}
\newcommand*{\axT}{\textbf{T}}
\newcommand*{\axFour}{$\mathbf{4}$}
\newcommand*{\axFive}{$\mathbf{5}$}
\newcommand*{\axD}{\textbf{D}}
\newcommand*{\axK}{\textbf{K}}
\newcommand*{\logic}[3]{\textbf{#1}$_{#2}^{\mathbf{#3}}$}
\newcommand*{\state}[2]{\ensuremath{(M_{\defemph{#1}},\defemph{#2})}}

\newcommand*{\sAG}{\ensuremath{\mathcal{AG}}}

\newcommand*{\sF}{\ensuremath{\mathcal{F}}}

\newcommand*{\func}[3]{#1: #2 \mapsto #3}
\newcommand*{\implies}{\ensuremath{\Rightarrow}}

\newcommand*{\bra}[1]{\ensuremath{\{#1\}}}
\newcommand*{\tuple}[1]{\ensuremath{\langle #1 \rangle}}

\newcommand*{\calB}{\ensuremath{\mathcal{B}}}

\newcommand*{\colorAgentSlide}[1]{\textcolor{Black}{#1}}
\newcommand*{\agentSlide}[1]{%
\ifstrequal{#1}{Charlie}{\colorAgentSlide{\texttt{#1}}}%
{\ifstrequal{#1}{Lucy}{\colorAgentSlide{\texttt{#1}}}%
{\ifstrequal{#1}{Snoopy}{\colorAgentSlide{\texttt{#1}}}%
{\ifstrequal{#1}{ag}{\colorAgentSlide{\texttt{#1}}}%
{\ifstrequal{#1}{ag_i}{\ensuremath{\colorAgentSlide{\mathtt{#1}}}}%
{\ifstrequal{#1}{ag_1}{\ensuremath{\colorAgentSlide{\mathtt{#1}}}}%
{\ifstrequal{#1}{ag_2}{\ensuremath{\colorAgentSlide{\mathtt{#1}}}}%
{\ifstrequal{#1}{AG}{\colorAgentSlide{\texttt{\lowercase{#1}}}}%
{\ifstrequal{#1}{A}{\colorAgentSlide{\texttt{#1}}}%
{\ifstrequal{#1}{B}{\colorAgentSlide{\texttt{#1}}}%
{\ifstrequal{#1}{C}{\colorAgentSlide{\texttt{#1}}}%
{\ifstrequal{#1}{a}{\colorAgentSlide{\uppercase{\texttt{#1}}}}%
{\ifstrequal{#1}{b}{\colorAgentSlide{\uppercase{\texttt{#1}}}}%
{\ifstrequal{#1}{c}{\colorAgentSlide{\uppercase{\texttt{#1}}}}%
{\ifstrequal{#1}{agent}{\colorAgentSlide{\texttt{#1}}}%
{\ifstrequal{#1}{agents}{\colorAgentSlide{\texttt{#1}}}{??%
}}}}}}}}}}}}}}}}%
}

\newcommand{\showCILC}[2]{%
	\ifstrequal{#1}{true}{#2}{}}

\theoremstyle{plain}

\theoremstyle{definition}
\newtheorem{definition}{Definition}[section]
\newtheorem{example}{Example}[section]
\title{Design of a Solver for \\ Multi-Agent Epistemic Planning}
\author{Francesco Fabiano
\institute{Dipartimento DMIF, Universit\`a di Udine, Udine, Italy}
\email{francesco.fabiano@uniud.it}}

\begin{document}
%\includepdf[pages=-]{vita/vita-ICLP.pdf}
%\includepdf[pages=-]{letters/dovier-Letter.pdf}
%\includepdf[pages=-]{letters/pontelli-Letter.pdf}
%\setcounter{page}{1}

	\maketitle              % typeset the header of the contribution
\begin{abstract}
	As the interest in Artificial Intelligence continues to grow it is becoming more and more important to investigate
	formalization and tools that allow us to exploit logic to reason about the world.
	In particular, given the increasing number of multi-agents systems that could benefit from
	techniques of automated reasoning, exploring new ways to define not only the world's status but
	also the agents' information is constantly growing in importance.
	This type of reasoning, \ie about agents' perception of the world and also about agents' knowledge of her and others' knowledge,
	is referred to as \emph{epistemic reasoning}.

	In our work we will try to formalize this concept, expressed through \emph{epistemic logic}, for dynamic domains.
	In particular we will attempt to define a new action-based language for \emph{multi-agent epistemic planning} and
	to implement an epistemic planner based on it.
	This solver should provide a tool flexible enough to be able to reason on different domains, \eg economy, security, justice and politics, where
	reasoning about others' beliefs could lead to winning strategies or help in changing a group of agents' view of the world.
	\\~\\{\bf Keywords:}
	\begin{enumerate*}[label=\hspace*{3pt}$\mathbf{\cdot}$\hspace*{3pt}]
		\item[] Epistemic reasoning
		\item Planning
		\item Multi-agent
		\item Action languages
		\item Heuristics
		\item[] Information change
		\item Knowledge representation
	\end{enumerate*}
\end{abstract}

\section{Introduction}
The proliferation of agent-based and IoT technologies has enabled the development of novel applications involving hundreds of agents.
Considering that self-driving cars and other autonomous devices that can control several aspects of our daily life are going to be available en mass in just a few years it will not be long until massive systems of autonomous agents, each acting upon its own knowledge and beliefs to achieve its own (or group) goals, become available and widely deployed.

To maximize the potentials of such autonomous systems, \emph{multi-agent planning} and scheduling research \cite{dovier1,de2009introduction,de2003resource, allen2009complexity, lipovetzky2017best, richter2010lama} will need to keep pace. Moreover creating a plan for multiple agents to achieve a goal will need to take into consideration agents' knowledge and beliefs, to account for aspects like trust, dishonesty, deception, and incomplete knowledge.
The planning problem in this new setting is referred to as \emph{epistemic planning} in the literature; that is epistemic planners are not only interested in the state of the world but also in the knowledge or beliefs of the agents.

Nevertheless, reasoning about knowledge and beliefs is not as direct as reasoning on the \textquotedblleft physical" state of the world. That is because expressing, for example, belief relations between a group of agents often implies to consider \emph{nested} and \emph{group} beliefs that are not easily extracted from the state description by a human reader. For this reasons it is necessary to develop a complete and accessible action language to model multi-agent epistemic domains~\cite{baral2015action} and to advance also in the study of epistemic solvers~\cite{le2018efp,bolander2011epistemic,wan2015complete,muise2015planning,huang2017general}.

In our research we are exploring the epistemic planning problem with particular focus on formalizing an updated version of the action language proposed in~\cite{baral2015action} and on developing a multi-agent epistemic solver flexible enough to be used in several scenarios such as: economy, security, justice and politics.
During this study we also intent to confront problems related to \emph{epistemic dynamic logic} such as:
 \begin{enumerate*}[label=\roman*)]
	\item the complexity of the plan existence problem (as explored in~\cite{bolander2015complexity}); and
	\item the use of alternative or implicit data structures that would help in representing epistemic states more efficiently.
\end{enumerate*}
\section{Background}\label{sec:background}
\subsection{Dynamic Epistemic Logic}
Logicians have always been interested in describing \emph{the state of the world} through formalism that would allow  to reason on the world with logic itself.
This interest has lead, among other things, to the formalization of the well-known planning problem~\cite{modernApproach} and to the introduction of several \emph{modal logics}~\cite{van2007dynamic,Chagrov1997,smullyan2012first} used to describe different types of domains.
The differences between these logics are not merely syntactical but they carry implications in both expressiveness and complexity.
Let us take for example, without going into details, the boolean \emph{propostional logic} and the \emph{linear temporal logic} (LTL).
The first one, being one of the simplest logics, is mostly used to encode the world as a series of facts that can be true or not and,
therefore, allows to \textquotedblleft reduce" properties of the domain to \emph{boolean formulae}.
The latter instead, even if it is based on propositional logic, introduces new \emph{modal operators} that allow to reason also about time (with a little abuse of the term).
The absence of these operators in the first one makes propositional logic, adopted to represent problems such as \textit{n-queens} or \textit{circuit-SAT}, not expressive enough to encode problems, \eg \emph{timeline-based planning problem}~\cite{DBLP:journals/corr/abs-1902-06123}, that LTL can deal with.
So, in general, we have that different logics have diverse operators and therefore are suitable for different type of reasoning on the world.
%It is clear that the two logics allow diverse types of reasoning and bring with them different implications in terms of expressiveness and complexity.

Nevertheless, even if different, both the two logics introduced above are limited in reasoning only on the state of world, \ie on its \textquotedblleft physical" properties and on their changes.
\emph{Dynamic Epistemic Logic} (\del), on the other hand, is used to reason not only on the state of the world but also on \emph{information change}.
As said in~\cite{van2007dynamic} \emph{information} is something that is relative to a subject who has a certain perspective on the world, called an \emph{agent}, and that is meaningful as a whole, not just loose bits and pieces. This makes us call it \emph{knowledge} and, to a lesser extent, \emph{belief}.
The idea behind \del\ is, therefore, to have a formalization that allows to reason on dynamic domains where, not only the state world is taken into consideration, but also the knowledge/beliefs that the agents have about the world and about the knowledge/beliefs of each other are considered.

\subsubsection{Epistemic Logic}
Dynamic Epistemic Logic is, clearly, connected to \emph{epistemic logic}: that is the logic that allows to reason on the knowledge/belief of agents in static domains.
This logic is based on two main concepts:
\begin{enumerate*}[label=\roman*)]
\item \emph{Kripke structures}, a data structure that is widely use in literature~\cite{fagin1994reasoning,baral2015action,van2007dynamic} to model its semantics; and
\item \emph{belief formulae}, a type of formula that takes into consideration epistemic operators and is used to represent the knowledge/beliefs of the agents.
\end{enumerate*}
As it is beyond the scope of this work to give an exhaustive introduction on epistemic logic we will provide only the fundamental definitions and intuitions; the reader who has interest in a more detailed description can refer to~\cite{fagin1994reasoning}.

Let \sAG\ be a set of  agents and let \sF\ be a set of propositional variables, called \emph{fluents}.
We have that each \emph{world} is described by a
subset of elements of \sF\ (intuitively, those that are \textquotedblleft true" in the world).
% We have that a state is represented by a consistent collection of worlds.
Moreover in epistemic logic each agent $\agent{ag} \in \sAG$ is associated with an epistemic modal operator \b{ag} %\footnote{Where $\mathbf{B}$ stands for knowledge or belief depending on the domain specification.}
that intuitively represents the knowledge/belief of \agent{ag}.
Finally, epistemic \emph{group operators} \eAlpha\ and \cAlpha\ are also introduced in epistemic logic. Intuitively \eAlpha\ and \cAlpha\ represent the knowledge/belief of a group of agents $\alpha$ and
the \emph{\ck/belief} of $\alpha$ respectively.
To be more precise, as in~\cite{baral2015action}, we have that:
\begin{definition}
	A \emph{fluent formula} is a propositional formula built using the propositional
	variables in \sF\ and the traditional propositional operators
	$\wedge,\vee,\implies,\neg$. We will use $\top$ and $\bot$ to indicate
	\emph{True} and \emph{False}, respectively.
	%\end{d}
	%\begin{definition}[fluent atom and literal]
	A \emph{fluent atom} is a formula composed by just an element $\defemph{f} \in \sF$, instead
	a \emph{fluent literal} is either a fluent atom $\defemph{f} \in \sF$ or its negation $\neg \defemph{f}$. During this work we will refer to fluent literals simply as \emph{fluents}.
\end{definition}

\begin{definition}
	A \emph{belief formula} is defined as follow:
	\begin{itemize}
		\item A fluent formula is a belief formula;
		\item let $\varphi$ be belief formula and $\agent{ag} \in \sAG$, then  $\bB{ag}{\varphi}$ is a belief
		      formula;
		\item let $\varphi_1, \varphi_2$ and $\varphi_3$ be belief formulae,
		      then $\neg \varphi_3$ and $\varphi_1 \,\mathtt{op}\, \varphi_2$ are belief
		      formulae, where $\mathtt{op} \in \bra{\wedge,\vee, \implies}$;
		\item all the formulae of the form \eAlpha{\varphi} or \cAlpha{\varphi}
		      are belief formulae, where $\varphi$ is itself a belief formula and
		      $\emptyset \neq \alpha \subseteq \sAG$.
	\end{itemize}
\end{definition}
From now on we will denote with \lagC\ the language of the belief formulae over the
sets $\sF$ and $\sAG$ and with \lag\ as the language over beliefs formulae that
does not allow the use of   \C.
%In~\cite{fagin1994reasoning}, it is pointed out how these two languages differ
%in expressiveness and complexity.

\begin{example}
	Let us consider the formula $\bB{ag_1}{\bB{ag_2}{\varphi}}$. This formula expresses that
	the agent \agent{ag_1} believes that the agent \agent{ag_2} believes that $\varphi$ is true, instead,
	$\bB{ag_1}\neg \varphi$ expresses that the agent \agent{ag_1} believes that $\varphi$ is false.
\end{example}
As mentioned above the classical way of providing a semantics for the language of epistemic logic is in terms
of \emph{pointed Kripke structure}~\cite{Kripke1963-KRISCO}. More formally:
%$\mAL$ use the Kripke structures to reasoning about the environment

\begin{definition}%[Kripke structure]
	A \emph{Kripke structure} is a tuple \tuple{S, \pi, \brel{1},$\dots$ , \brel{n}}, such that:
	\begin{itemize}
		\item S is a set of worlds;
		\item $\func{\pi}{S}{2^{\sF}}$ is a function that associates an interpretation
		of \sF\ to each element of S; %(namely the subset of fluents $\top$ in each world);
		\item for $1 \leq \defemph{i} \leq \defemph{n}$, $\brel{i} \subseteq S \times S$  is a binary relation over S.
	\end{itemize}
\end{definition}
\begin{definition}%[Kripke structure]
A \emph{pointed Kripke structure} is a pair \state{}{s} where $M$ is a Kripke structure
as defined above, and $\defemph{s} \in S$, where $\defemph{s}$ represents the real world.
\end{definition}
Following the notation of~\cite{baral2015action}, we will indicate with
$M[S], M[\pi],$ and $M[\defemph{i}]$ the components $S,\pi$, and $\brel{i}$ of $M$,
respectively.
Intuitively  $M[S]$ captures all the worlds that the agents believe to be possible and $M[\defemph{i}]$ encodes the beliefs of each agent. More formally the semantics on pointed Kripke structure is as follows:

\begin{definition}%[entailment w.r.t. a Kripke structure]
	Given the belief formulae
	$\varphi,\varphi_{1},\varphi_{2}$, an agent \agent{ag_i}, a group of agents $\alpha$, a pointed Kripke structure ($M = \tuple{S, \pi, \brel{1}, ..., \brel{n}}$, \defemph{s}):
	\begin{enumerate}[label= \emph{(}\roman*\emph{)}]
		\item $\state{}{s} \models \varphi$ if $\varphi$ is a fluent formula and $\pi(\defemph{s})
		\models \varphi$;
		\item $\state{}{s} \models \bB{ag_i}{\varphi}$ if for each \defemph{t} such that
		$\defemph{(s,t)} \in \brel{i}$ it holds that $ \state{}{t} \models \varphi$;
		%\item $\state{}{s} \models \neg \varphi$ if $\state{}{s} \not\models \varphi$;
	%	\item $\state{}{s} \models \varphi_1 \vee \varphi_2$ if $\state{}{s}\models
		%\varphi_1$ or $\state{}{s}\models \varphi_2$;
		%\item $\state{}{s} \models \varphi_1 \wedge \varphi_2$ if $\state{}{s}\models
	%	\varphi_1$ and $\state{}{s}\models \varphi_2$;
		\item $\state{}{s} \models \eAlpha{\varphi}$ if $\state{}{s} \models \bB{ag_i}{\varphi}$ for all \agent{ag_i} $\in \alpha$;
		\item $\state{}{s} \models \cAlpha{\varphi}$ if
		$\state{}{s} \models \eAlphaIter{k}{\varphi}$ for every
		$k\geq0$, where $\eAlphaIter{0}{\varphi} = \varphi$ and
		$\eAlphaIter{k+1}{\varphi} =\eAlpha{(\eAlphaIter{k}{\varphi})}$;
		\item the semantics of the traditional propositional operators is as usual.
	\end{enumerate}
\end{definition}

\subsubsection{Knowledge or Belief}

	As pointed out in the previous paragraph the modal operator $\bB{ag}{}$ represents the worlds' relation in a Kripke structure 
	and, as expected,	different relations' properties imply different meaning for $\bB{ag}{}$.
	In particular in this work we are interested in representing the knowledge or the beliefs of the agents.
	The problem of formalizing these two concepts has been studied in depth bringing to an accepted formalization for both knowledge and beliefs.
	In fact we have that when a relation\footnote{In our case the relation between the world in a Kripke structure.} respects all the axioms presented in Table~\ref{tab:axioms} is called an \emph{\textbf{S5}} relation and it encodes the concept of knowledge while when it encodes all the axioms but \axT\ it characterizes the concept of belief.
	Following these characterization we will refer to knowledge and belief as \textbf{S5} and \textbf{KD45} logic respectively.

	\begin{table}[h]
		\centering
		\begin{tabular}{||c||c||}
			\hhline{|t:=:t:=:t|}
			\multicolumn{1}{||c||}{\phantom{...}\textbf{Property of $\brel{}$}\phantom{...}}
			& \multicolumn{1}{c||}{\phantom{...}\textbf{Axiom}\phantom{...}}\\
			\hhline{|:=::=:|}
			\multicolumn{1}{||l||}{$\calB_{\mathtt{i}}\varphi  \implies \varphi$}
			& \multicolumn{1}{c||}{\axT}\\
			\hhline{||-||-||}
			\multicolumn{1}{||l||}{$\calB_{\mathtt{i}}\varphi  \implies \calB_{\mathtt{i}}\calB_{\mathtt{i}}\varphi$}
			& \multicolumn{1}{c||}{\axFour}\\
			\hhline{||-||-||}
			\multicolumn{1}{||l||}{$\neg \calB_{\mathtt{i}}\varphi  \implies \calB_{\mathtt{i}}\neg \calB_{\mathtt{i}}\varphi$}
			& \multicolumn{1}{c||}{\axFive}\\
			\hhline{||-||-||}
			\multicolumn{1}{||l||}{$\neg \calB_{\mathtt{i}} \bot$}
			& \multicolumn{1}{c||}{\axD}\\
			\hhline{||-||-||}
			\multicolumn{1}{||l||}{$(\calB_{\mathtt{i}}\varphi \wedge \calB_{\mathtt{i}}(\varphi \implies \psi)) \implies  \calB_{\mathtt{i}}\psi$}
			& \multicolumn{1}{c||}{\axK}\\
			\hhline{|b:=:b:=:b|}
		\end{tabular}
		\caption{\label{tab:axioms}Knowledge and beliefs axioms.~\cite{fagin1994reasoning}.}
	\end{table}
Intuitively the difference between the two logics is that an agent cannot \textit{know} something that is not true in \textbf{S5} but she can \textit{believe} it in \textbf{KD45}.
	As this introduction is not supposed to explore in depth this topic we will not go into further detail and we address the interested reader to~\cite{fagin1994reasoning}.
\subsubsection{Complexity Overview}
Finally, as last note on DEL we will present two short, but informative, tables that summarize the complexity results in the epistemic logic and in the epistemic planning fields
(Table~\ref{tab:comlog} and Table~\ref{tab:complan} respectively).
The information presented in these tables serve to provide the reader with a general idea on \textquotedblleft how hard" the problem of reasoning on information change is.
\begin{table}[h]
\raisebox{0.3cm}{\begin{subtable}[t]{0.48\textwidth}
	\centering
	\begin{tabular}{||>{\centering\arraybackslash}m{1.4in}|>{\centering\arraybackslash}m{1.0in}||}
		\hhline{|t:==:t|}
		\multicolumn{1}{||c|}{\phantom{...}{\emph{\textbf{SAT}} Complexity}\phantom{...}}
		& \multicolumn{1}{c||}{\phantom{.}{Epistemic logic}\phantom{.}}\\
		\hhline{|:==:|}
		%\hhline{||-|-||}
		NP-complete&\logic{S5}{1}{}, \logic{KD45}{1}{}\\
		\hhline{||-|-||}
		PSPACE-complete&\logic{S5}{n}{}, \logic{KD45}{n}{}\newline with $n\geq 2$\\
		\hhline{||-|-||}
		EXPTIME-complete &\logic{S5}{n}{C}, \logic{KD45}{n}{C}\newline with $n\geq 2$\\
		\hhline{|b:==:b|}
	\end{tabular}
	\caption{\label{tab:comlog}Complexity of the satisfiability problem w.r.t. knowledge and beliefs logics~\cite{fagin1994reasoning}.}
\end{subtable}%
}
\hfill
\begin{subtable}{0.48\textwidth}
	{\small
	\centering
	\begin{tabular}{||>{\centering\arraybackslash}m{1.7in} |>{\centering\arraybackslash}m{1.2in} ||}
		\hhline{|t:==:t|}
		{Action type}&
		{Plan Existence Complexity} \\
		\hhline{|:==:|}
		{Non factual with propositional preconditions}&
		\textsc{EXPSPACE}\\
		\hhline{||-|-||}
		{Factual with propositional preconditions} &
		\textsc{NON-elementary}\\
		\hhline{||-|-||}
		{Factual with epistemic preconditions} &
		\textsc{Undecidable} \\
		\hhline{|b:==:b|}
	\end{tabular}
	\caption{\label{tab:complan} Complexity of the \emph{plan existence problem}~\cite{bolander2015complexity}.}
}
\end{subtable}
\caption{Complexity results in dynamic epistemic logic.}
\end{table}

\subsection{Multi-Agent Epistemic Planning}
\emph{Epistemic planning}~\cite{bolander2011epistemic} refers to the generation of plans for multiple agents to achieve goals which can refer to the state of the world, the beliefs of agents, and/or the knowledge of agents. It has recently attracted the attention of researchers from various communities such as planning, dynamic epistemic logic, and knowledge representation.

With the introduction of the classical planning problem, in the early days of artificial intelligence, several action languages (e.g., $\mathcal{A}$, $\mathcal{B}$, and $\mathcal{C}$) have been developed~\cite{gelfond1998action} and have also provided the foundations for several successful approaches to automated planning.
However, the main focus of these research efforts has been about reasoning within single agent domains. 
In single agent domains reasoning about information change mainly involves reasoning about what the agent knows about the world and how she can manipulate it to reach particular states.
In multi-agent domains, on the other hand, an agent's action may change the world, other agents' knowledge about it and their knowledge about other agents' knowledge about the world. Similarly, goals of an agent in a multi-agent domain may involve manipulating the knowledge of other agents---in particular, this may concern not just their knowledge about the world, but also their knowledge about other agents' knowledge about the world.

As said before, epistemic planning is not only interested in the state of the world but also in the beliefs and the knowledge of the agents. Although there is a large body of research on multi-agent planning~\cite{de2003resource, de2009introduction, allen2009complexity,  goldman2004decentralized, guestrin2002multiagent}, very few efforts address the above mentioned aspects of multi-agent domains, which pose a number of new research challenges in representing and reasoning about actions and change.

Due to its complexity, the majority of search based epistemic planners (e.g.,~\cite{crosby2014single,engesser2017cooperative,kominis2015beliefs, kominis2017multiagent, huang2017general,muise2015planning, wan2015complete}) impose certain restrictions, such as the finiteness of the levels of nested beliefs. 
Such restrictions permit, for instance, in~\cite{muise2015planning, huang2017general, wan2015complete} to solve the problem by translating it into classical planning.
%Issues and research directions related to epistemic planning have been summarized nicely in the report of the 2017 Dagstuhl's meeting~\cite{baral2018dagstuhl}.

On the other hand, to the best of our knowledge, only few systems~\cite{le2018efp,liu2018multi} can reason about epistemic knowledge in multi-agent domains without these limitations.
In particular~\cite{le2018efp} and the language \mAL, that is the language that~\cite{le2018efp} implements, are the starting points of this research work.
As we will explain in the following sections the main objectives of this thesis are:
\begin{enumerate*}[label=\roman*)]
	\item to formalize a complete and flexible epistemic action-based language;
	\item to study alternative representations for epistemic models;
	\item to implement a competitive planner to reason on information change; and
	\item to explore the concept of heuristics in epistemic planning.
\end{enumerate*} 
Given the amount of information to properly introduce \mAL\ and the concept of \emph{event update semantics} we redirect the reader who is not familiar with these two topics to~\cite{baral2015action} for a clear and complete explanation.
\section{Goals of the research}\label{sec:goals}
As already pointed in the previous sections our work tries to explore in depth the multi-agent epistemic planning problem.
In particular this research aims to tackle the problem from several points of view, with the ultimate goal to provide a design (and an implementation) of an 
efficient \emph{epistemic planner} that can reason on the full extent of \logic{S5}{n}{C} and \logic{KD45}{n}{C} with $n \geq 1$.

To reach this main objective we are investigating multi-agent epistemic planning from different levels.
Following the the major sub-goals of our work will be presented.
\begin{itemize}[leftmargin=*]
	\item To formalize an \textbf{action-based language} that would be used to express epistemic domains.
	This new language will use~\cite{baral2015action} as base.
	Our design of a language will be mainly interested in dealing with:
	\begin{enumerate*}[label=\roman*)]
		\item \textquotedblleft \emph{false beliefs}", \ie the situation when an agent has false beliefs about the world and tries to modify the world itself.
		Solutions to this problem may be found in literature, \eg~\cite{le2018efp}, but are somewhat unsatisfactory.
		\item We will also be interested in providing a formalization for the concepts of \emph{trust}, \emph{lies} and \emph{deception}.
		\item Finally, we would also like to increase the expressiveness of \mAL, \eg by adding the possibility to perform announcements of belief formulae.
	\end{enumerate*}
	 
	 \item To study \textbf{alternative data structures} for epistemic states.
	 This can be done on two different levels:
	 \begin{itemize}[leftmargin=*]
	 	\item Introducing completely new data structures exploiting, for example, concepts from the widely studied knowledge representation~\cite{Brachman:1985:RKR:577033} and graph theory~\cite{Omodeo20171} literature.
	 	\item Another interesting approach, following what is done in the \emph{model checking} community, is to represent the states through symbolic data structures.
	 	This will also imply the introduction of a transition function that can deal with symbolic state representation.
	 \end{itemize}
 
 	 \item As last, and maybe most substantial, contribution of the thesis we would like to design and implement a flexible and complete \textbf{epistemic solver}.
 	 Whilst the formalization of the language and the introduction of new data structures will surely help in reducing the search time, the inherent complexity of DEL will always be a strong
 	 limitation for epistemic planners. That is why our work will be focused on designing a solver that makes a strong use of heuristics. In fact, as shown in~\cite{le2018efp}, reasoning on the full extent of \logic{KD45}{n}{C} at the moment is only possible when the length of the problems is really short (w.r.t. to the classical planning ones). That is why we believe that to provide a significant upgrade in performance it is necessary to introduce heuristics in this setting.
 	 As final note on the planner it is important to clarify that our solver could also be useful when exploited without heuristics.
 	 For example it could be used to reason on small domain where the information about the world is so intricate that would impossible for humans to reason about it.
 	 Providing examples of the various scenarios where the planner could be used, even with different configurations, will also be taken into consideration during our research. 
\end{itemize}
\section{Status of the research}\label{sec:current}
In this section we quickly present the work done on the goals introduced previously.
Given the early stage of the research the results are often not completely demonstrated or fully tested;
nevertheless we are positive to accomplish that in the next future.
We will now provide, without tedious technical details, the status of the research of each one of the three sub-goals presented in Section~\ref{sec:goals};
the reader who is interested in a more complete introduction to these is mainly addressed to~\cite{le2018efp,cilc19Epistemic}.

\begin{itemize}[leftmargin=*]
	\item \textbf{Language formalization:} we are currently investigating different lines of work that consider
	      the description of a new epistemic action-based language from diverse points of view.
	      First of all let us stress once again that this new language will be strongly based on \mAL~\cite{baral2015action}.
	      \begin{itemize}[leftmargin=*]
		      \item In respect to the \textquotedblleft false belief" problem we sketched a solution that involves to apply some small changes to the transition function of \mAL.
		            Intuitively we think that by allowing epistemic actions (\ie \emph{sensing} and \emph{announcement}) to modify not only the edges of the state but also its nodes
		            could provide a clean solution to this problem.
		      \item On the other hand we introduced in \mAL\ the concept of \emph{trust} as a globally visible relation between agents.
		            From this concept we formalized, thanks to the introduction of two new \emph{events} in the announcement transition function, the \emph{(un)trustworthy announcement} that
		            takes into consideration the trust that the listener has w.r.t. the announcer.
		      \item Starting from the concept of (un)trustworthy announcements we were also able to expand\footnote{Correctness still to be proven.} the language from being able to characterize only
		            fluent formulae announcement to being able to announce belief formulae as well.
	      \end{itemize}

	\item \textbf{Alternative data structures:} as already said  Kripke structure is the classical way to provide semantics for epistemic logic but this does not mean that is the only one.
	      In particular in~\cite{cilc19Epistemic} we introduced an extension of \mAL, called \ourL, that bases the state representation on \emph{\posS}, a data structure introduced in~\cite{Gerbrandy1997} and derived from \emph{\nwf\ sets} theory. As shown in the paper, using these structures allows to have a more accurate concept of epistemic state equality.
	      Moreover, being \posS\ based on sets, we think that some aspects of epistemic planning, \eg the group operators, could be reduced to set operations providing a cleaner semantics and a faster implementation.

	\item \textbf{Epistemic Solver:} as main objective of our research we hope to provide a multi-agent epistemic planner flexible enough to be used in real world scenarios.
	      As starting point we took the solver presented in~\cite{le2018efp} and we reformatted it. At this point of our study we have a modular C\texttt{++} solver that:
	      \begin{itemize}[leftmargin=*]
		      \item encodes the language \mAL\ and is as powerful as the ones presented in~\cite{le2018efp};
		      \item exploits the dynamic programming paradigm to encode more efficiently the search-space;
		      \item has a more clean transition function that prunes, when necessary, useless nodes of the Kripke structure to reduce the memory overhead;
		      \item allows parametric state representation. Thanks to this it is now only necessary to describe a new state representation, \eg possibilities,	in the programming language
		            and the solver will be able to reason on it. This feature will permit to have a planner with several options that could be used differently depending on the domain.
	      \end{itemize}

	\item \textbf{Heuristics:} with~\cite{le2018efp} we started to explore the concept of heuristics in epistemic planning.
	      At this point of the our study we did not implement new heuristics yet but we are constantly exploring new options.
	      Few informal ideas that we have had are:
	      \begin{itemize}[leftmargin=*]
		      \item to provide a normal form for belief formulae to introduce \emph{mutex} in the \emph{Epistemic Planning-Graph};
		      \item to solve epistemic problems as an instance of classical planning (with limited nested knowledge) to score the e-states;
		      \item to use techniques of local search where, as parameters to optimize, we take into consideration the knowledge or the ignorance of a group of agents;
		      \item and finally, to exploit more classical approaches, \eg the Monte Carlo strategy, to try to reduce the search-space.
	      \end{itemize}
\end{itemize}

In this brief section we defined what we think are the main aspects of our research.
Nevertheless there are other directions that our work can follows as epistemic planning is a relatively new field of study and needs
to be explored from several points of view.
In the following a short and summarizing list of alternative research directions is given.
\begin{itemize}[leftmargin=*]
	\item In~\cite{son2014finitary} the computation of the \emph{initial state} in \logic{S5}{}{} is addressed but it is still an open question how to
	      allow the same type of construction in \logic{KD45}{}{}. We formalized a possible solution for this problem but is yet to be verified thoroughly.
	\item Due to their high computational power, ASP solvers could be exploited to solve the epistemic planning problem or
	      at least some of its components (as in~\cite{DBLP:conf/iclp/SonPGB16}).
	\item On the same line of the previous point there is the possibility to exploit well known techniques from the model checking field to deal with the
	      \ck\ entailment. This operator is, in fact, strongly connected to the notion of \emph{fixed point}.
	\item Finally, an important feature of classical planning problems that has not be formalized yet for the epistemic ones are \emph{static laws}.
	      Static laws could play a really important role from both the expressiveness and flexibility points of view.
\end{itemize}
%
%
%
%\section{Preliminary results} \label{sec:results}
%\input{ch5-results}
%
%
%
\section{Open issues} \label{sec:issues}
Multi-agent planning and DEL are widely studied areas but their combination is relatively new and less explored.
In addition to that multi-agent epistemic planning, given its heavy formalism and
inherent complexity, it is not a really \textquotedblleft accessible" field of study and therefore not easy to be tackled.
In the following we list the main problems that we are facing in working on this project.

\begin{itemize}[leftmargin=*]
	\item First of all a small non-technical issue:
	      we noticed a substantial lack of real-world examples where epistemic planning can be adopted and that could help in responding to the question \textquotedblleft why using epistemic planners when the classical ones are extremely faster?"
	      We, ourselves, are trying to provide real-world scenarios that could exploit multi-agent epistemic planning but
	      we recognize that experts from other fields, \eg law or economics, could be very helpful in formalizing more attractive examples.

	\item Another problem is the one relative to heuristics. As we are not dealing only with world's properties we should
	      include in our heuristics also the concepts of belief (or knowledge). The main problem that we are facing when considering heuristics
	      is that is not always knowledge what an epistemic planner seeks but, potentially, different combinations of ignorance and knowledge for each agent.

	\item From the action-language point of view the problems reside in the complexity of its formalism.
	      While understanding every single detail of an epistemic planning language can be really challenging,
	      implementing or changing it is even harder. This is reflected into a general difficulty of the topic that
	      often discourage new researchers in investigating this field. A cleaner formalization of the epistemic planning
	      problem could help the community to grow.

	\item One last problem is the subjectivity of some epistemic-related concepts.
	      For example what is a lie? Or, again, when is an agent being deceptive?
	      Formalizing these concepts requires rules that may change from person to person or from situation to situation.
\end{itemize}

\section{Conclusions}
In this paper we presented the main concepts regarding our ideas
on how to develop a solver for multi-agent epistemic planning problems.
We identified three major sub-goals that, in our opinion,
are the ones that are the most significant for the epistemic planning
community and we provided some research directions that we intent to 
follow in the next future.
Moreover, we identified the major issues we are encountering while
developing our ideas that could be used as starting points for 
related works that could bring benefits to the entire community.
\section{Acknowledgements}
The work described in this paper has been partially supported
 by the Università di Udine PRID ENCASE project, and by
GNCS-INdAM 2017 and 2019 projects.

\bibliographystyle{eptcs}
\bibliography{utility/epistemicBib}

\begin{thebibliography}{10}
\providecommand{\bibitemdeclare}[2]{}
\providecommand{\surnamestart}{}
\providecommand{\surnameend}{}
\providecommand{\urlprefix}{Available at }
\providecommand{\url}[1]{\texttt{#1}}
\providecommand{\href}[2]{\texttt{#2}}
\providecommand{\urlalt}[2]{\href{#1}{#2}}
\providecommand{\doi}[1]{doi:\urlalt{http://dx.doi.org/#1}{#1}}
\providecommand{\bibinfo}[2]{#2}

\bibitemdeclare{inproceedings}{allen2009complexity}
\bibitem{allen2009complexity}
\bibinfo{author}{Martin \surnamestart Allen\surnameend} \&
  \bibinfo{author}{Shlomo \surnamestart Zilberstein\surnameend}
  (\bibinfo{year}{2009}): \emph{\bibinfo{title}{Complexity of decentralized
  control: Special cases}}.
\newblock In: {\sl \bibinfo{booktitle}{Advances in Neural Information
  Processing Systems}}, pp. \bibinfo{pages}{19--27}.

\bibitemdeclare{article}{baral2015action}
\bibitem{baral2015action}
\bibinfo{author}{Chitta \surnamestart Baral\surnameend},
  \bibinfo{author}{Gregory \surnamestart Gelfond\surnameend},
  \bibinfo{author}{Enrico \surnamestart Pontelli\surnameend} \&
  \bibinfo{author}{Tran~Cao \surnamestart Son\surnameend}
  (\bibinfo{year}{2015}): \emph{\bibinfo{title}{An Action Language for
  Multi-Agent Domains: Foundations}}.
\newblock {\sl \bibinfo{journal}{CoRR}} \bibinfo{volume}{abs/1511.01960}.
\newblock \urlprefix\url{http://arxiv.org/abs/1511.01960}.

\bibitemdeclare{conference}{bolander2015complexity}
\bibitem{bolander2015complexity}
\bibinfo{author}{T.~\surnamestart Bolander\surnameend}, \bibinfo{author}{M.H.
  \surnamestart Jensen\surnameend} \& \bibinfo{author}{F.~\surnamestart
  Schwarzentruber\surnameend} (\bibinfo{year}{2015}):
  \emph{\bibinfo{title}{Complexity results in epistemic planning}}.
\newblock In: {\sl \bibinfo{booktitle}{{IJCAI} International Joint Conference
  on Artificial Intelligence}}, \bibinfo{volume}{2015-January}, pp.
  \bibinfo{pages}{2791--2797}.

\bibitemdeclare{article}{bolander2011epistemic}
\bibitem{bolander2011epistemic}
\bibinfo{author}{Thomas \surnamestart Bolander\surnameend} \&
  \bibinfo{author}{Mikkel~Birkegaard \surnamestart Andersen\surnameend}
  (\bibinfo{year}{2011}): \emph{\bibinfo{title}{Epistemic planning for
  single-and multi-agent systems}}.
\newblock {\sl \bibinfo{journal}{Journal of Applied Non-Classical Logics}}
  \bibinfo{volume}{21}(\bibinfo{number}{1}), pp. \bibinfo{pages}{9--34},
  \doi{10.1016/0010-0277(83)90004-5}.

\bibitemdeclare{book}{Brachman:1985:RKR:577033}
\bibitem{Brachman:1985:RKR:577033}
\bibinfo{editor}{Ronald~J. \surnamestart Brachman\surnameend} \&
  \bibinfo{editor}{Hector~J. \surnamestart Levesque\surnameend}, editors
  (\bibinfo{year}{1985}): \emph{\bibinfo{title}{Readings in Knowledge
  Representation}}.
\newblock \bibinfo{publisher}{Morgan Kaufmann Publishers Inc.},
  \bibinfo{address}{San Francisco, CA, USA}.

\bibitemdeclare{book}{Chagrov1997}
\bibitem{Chagrov1997}
\bibinfo{author}{Alexander \surnamestart Chagrov\surnameend}
  (\bibinfo{year}{1997}): \emph{\bibinfo{title}{Modal Logic}}.
\newblock \bibinfo{publisher}{Oxford University Press}.

\bibitemdeclare{inproceedings}{crosby2014single}
\bibitem{crosby2014single}
\bibinfo{author}{Matthew \surnamestart Crosby\surnameend},
  \bibinfo{author}{Anders \surnamestart Jonsson\surnameend} \&
  \bibinfo{author}{Michael \surnamestart Rovatsos\surnameend}
  (\bibinfo{year}{2014}): \emph{\bibinfo{title}{A single-agent approach to
  multiagent planning}}.
\newblock In: {\sl \bibinfo{booktitle}{Proceedings of the Twenty-first European
  Conference on Artificial Intelligence}}, \bibinfo{organization}{IOS Press},
  pp. \bibinfo{pages}{237--242}.

\bibitemdeclare{article}{de2003resource}
\bibitem{de2003resource}
\bibinfo{author}{Mathijs \surnamestart De~Weerdt\surnameend},
  \bibinfo{author}{Andr{\'e} \surnamestart Bos\surnameend},
  \bibinfo{author}{Hans \surnamestart Tonino\surnameend} \&
  \bibinfo{author}{Cees \surnamestart Witteveen\surnameend}
  (\bibinfo{year}{2003}): \emph{\bibinfo{title}{A resource logic for
  multi-agent plan merging}}.
\newblock {\sl \bibinfo{journal}{Annals of Mathematics and Artificial
  Intelligence}} \bibinfo{volume}{37}(\bibinfo{number}{1-2}), pp.
  \bibinfo{pages}{93--130}, \doi{10.1023/A:1020236119243}.

\bibitemdeclare{article}{de2009introduction}
\bibitem{de2009introduction}
\bibinfo{author}{Mathijs \surnamestart De~Weerdt\surnameend} \&
  \bibinfo{author}{Brad \surnamestart Clement\surnameend}
  (\bibinfo{year}{2009}): \emph{\bibinfo{title}{Introduction to planning in
  multiagent systems}}.
\newblock {\sl \bibinfo{journal}{Multiagent and Grid Systems}}
  \bibinfo{volume}{5}(\bibinfo{number}{4}), pp. \bibinfo{pages}{345--355},
  \doi{10.3233/MGS-2009-0133}.

\bibitemdeclare{article}{dovier1}
\bibitem{dovier1}
\bibinfo{author}{Agostino \surnamestart Dovier\surnameend},
  \bibinfo{author}{Andrea \surnamestart Formisano\surnameend} \&
  \bibinfo{author}{Enrico \surnamestart Pontelli\surnameend}
  (\bibinfo{year}{2013}): \emph{\bibinfo{title}{{Autonomous agents
  coordination: Action languages meet CLP() and Linda}}}.
\newblock {\sl \bibinfo{journal}{Theory and Practice of Logic Programming}}
  \bibinfo{volume}{13}(\bibinfo{number}{2}), pp. \bibinfo{pages}{149--173},
  \doi{10.1016/S0004-3702(00)00031-X}.

\bibitemdeclare{article}{engesser2017cooperative}
\bibitem{engesser2017cooperative}
\bibinfo{author}{Thorsten \surnamestart Engesser\surnameend},
  \bibinfo{author}{Thomas \surnamestart Bolander\surnameend},
  \bibinfo{author}{Robert \surnamestart Mattm{\"u}ller\surnameend} \&
  \bibinfo{author}{Bernhard \surnamestart Nebel\surnameend}
  (\bibinfo{year}{2017}): \emph{\bibinfo{title}{Cooperative epistemic
  multi-agent planning for implicit coordination}}.
\newblock {\sl \bibinfo{journal}{arXiv preprint arXiv:1703.02196}}.

\bibitemdeclare{inproceedings}{cilc19Epistemic}
\bibitem{cilc19Epistemic}
\bibinfo{author}{Francesco \surnamestart Fabiano\surnameend},
  \bibinfo{author}{Idriss \surnamestart Riouak\surnameend},
  \bibinfo{author}{Agostino \surnamestart Dovier\surnameend} \&
  \bibinfo{author}{Enrico \surnamestart Pontelli\surnameend}
  (\bibinfo{year}{2019}): \emph{\bibinfo{title}{Non-well-founded set based
  multi-agent epistemic action language}}.
\newblock In: {\sl \bibinfo{booktitle}{Proceedings of the 34rd Italian
  Conference on Computational Logic}}, {\sl \bibinfo{series}{{CEUR} Workshop
  Proceedings}} \bibinfo{volume}{2396}, \bibinfo{address}{Trieste, Italy}, pp.
  \bibinfo{pages}{242--259}.
\newblock \urlprefix\url{http://ceur-ws.org/Vol-2396/paper38.pdf}.

\bibitemdeclare{article}{fagin1994reasoning}
\bibitem{fagin1994reasoning}
\bibinfo{author}{Ronald \surnamestart Fagin\surnameend} \&
  \bibinfo{author}{Joseph~Y \surnamestart Halpern\surnameend}
  (\bibinfo{year}{1994}): \emph{\bibinfo{title}{Reasoning about knowledge and
  probability}}.
\newblock {\sl \bibinfo{journal}{Journal of the ACM (JACM)}}
  \bibinfo{volume}{41}(\bibinfo{number}{2}), pp. \bibinfo{pages}{340--367},
  \doi{10.1145/174652.174658}.

\bibitemdeclare{article}{gelfond1998action}
\bibitem{gelfond1998action}
\bibinfo{author}{Michael \surnamestart Gelfond\surnameend} \&
  \bibinfo{author}{Vladimir \surnamestart Lifschitz\surnameend}
  (\bibinfo{year}{1998}): \emph{\bibinfo{title}{Action Languages}}.
\newblock {\sl \bibinfo{journal}{Electron. Trans. Artif. Intell.}}
  \bibinfo{volume}{2}, pp. \bibinfo{pages}{193--210}.
\newblock \urlprefix\url{http://www.ep.liu.se/ej/etai/1998/007/}.

\bibitemdeclare{article}{Gerbrandy1997}
\bibitem{Gerbrandy1997}
\bibinfo{author}{J.~\surnamestart Gerbrandy\surnameend} \&
  \bibinfo{author}{W.~\surnamestart Groeneveld\surnameend}
  (\bibinfo{year}{1997}): \emph{\bibinfo{title}{Reasoning about information
  change}}.
\newblock {\sl \bibinfo{journal}{Journal of Logic, Language and Information}}
  \bibinfo{volume}{6}(\bibinfo{number}{2}), pp. \bibinfo{pages}{147--169},
  \doi{10.1023/A:1008222603071}.

\bibitemdeclare{article}{DBLP:journals/corr/abs-1902-06123}
\bibitem{DBLP:journals/corr/abs-1902-06123}
\bibinfo{author}{Nicola \surnamestart Gigante\surnameend}
  (\bibinfo{year}{2019}): \emph{\bibinfo{title}{Timeline-based planning:
  Expressiveness and Complexity}}.
\newblock {\sl \bibinfo{journal}{CoRR}} \bibinfo{volume}{abs/1902.06123}.

\bibitemdeclare{article}{goldman2004decentralized}
\bibitem{goldman2004decentralized}
\bibinfo{author}{Claudia~V \surnamestart Goldman\surnameend} \&
  \bibinfo{author}{Shlomo \surnamestart Zilberstein\surnameend}
  (\bibinfo{year}{2004}): \emph{\bibinfo{title}{Decentralized control of
  cooperative systems: Categorization and complexity analysis}}.
\newblock {\sl \bibinfo{journal}{J. Artif. Intell. Res.(JAIR)}}
  \bibinfo{volume}{22}, pp. \bibinfo{pages}{143--174}, \doi{10.1613/jair.1427}.

\bibitemdeclare{inproceedings}{guestrin2002multiagent}
\bibitem{guestrin2002multiagent}
\bibinfo{author}{Carlos \surnamestart Guestrin\surnameend},
  \bibinfo{author}{Daphne \surnamestart Koller\surnameend} \&
  \bibinfo{author}{Ronald \surnamestart Parr\surnameend}
  (\bibinfo{year}{2002}): \emph{\bibinfo{title}{Multiagent planning with
  factored MDPs}}.
\newblock In: {\sl \bibinfo{booktitle}{Advances in neural information
  processing systems}}, pp. \bibinfo{pages}{1523--1530}.

\bibitemdeclare{conference}{huang2017general}
\bibitem{huang2017general}
\bibinfo{author}{X.~\surnamestart Huang\surnameend},
  \bibinfo{author}{B.~\surnamestart Fang\surnameend},
  \bibinfo{author}{H.~\surnamestart Wan\surnameend} \&
  \bibinfo{author}{Y.~\surnamestart Liu\surnameend} (\bibinfo{year}{2017}):
  \emph{\bibinfo{title}{A general multi-agent epistemic planner based on
  higher-order belief change}}.
\newblock In: {\sl \bibinfo{booktitle}{{IJCAI} International Joint Conference
  on Artificial Intelligence}}, pp. \bibinfo{pages}{1093--1101}.

\bibitemdeclare{inproceedings}{kominis2015beliefs}
\bibitem{kominis2015beliefs}
\bibinfo{author}{Filippos \surnamestart Kominis\surnameend} \&
  \bibinfo{author}{Hector \surnamestart Geffner\surnameend}
  (\bibinfo{year}{2015}): \emph{\bibinfo{title}{Beliefs In Multiagent Planning:
  From One Agent to Many}}.
\newblock In: {\sl \bibinfo{booktitle}{Proceedings of the International
  Conference on Automated Planning and Scheduling, {ICAPS}}}, pp.
  \bibinfo{pages}{147--155}.

\bibitemdeclare{inproceedings}{kominis2017multiagent}
\bibitem{kominis2017multiagent}
\bibinfo{author}{Filippos \surnamestart Kominis\surnameend} \&
  \bibinfo{author}{Hector \surnamestart Geffner\surnameend}
  (\bibinfo{year}{2017}): \emph{\bibinfo{title}{Multiagent Online Planning with
  Nested Beliefs and Dialogue}}.
\newblock In: {\sl \bibinfo{booktitle}{Proceedings of the International
  Conference on Automated Planning and Scheduling, {ICAPS}}},
  \bibinfo{address}{Pittsburgh, Pennsylvania, USA}, pp.
  \bibinfo{pages}{186--194}.
\newblock
  \urlprefix\url{https://aaai.org/ocs/index.php/ICAPS/ICAPS17/paper/view/15748}.

\bibitemdeclare{article}{Kripke1963-KRISCO}
\bibitem{Kripke1963-KRISCO}
\bibinfo{author}{Saul~A. \surnamestart Kripke\surnameend}
  (\bibinfo{year}{1963}): \emph{\bibinfo{title}{Semantical Considerations on
  Modal Logic}}.
\newblock {\sl \bibinfo{journal}{Acta Philosophica Fennica}}
  \bibinfo{volume}{16}(\bibinfo{number}{1963}), pp. \bibinfo{pages}{83--94}.

\bibitemdeclare{inproceedings}{le2018efp}
\bibitem{le2018efp}
\bibinfo{author}{Tiep \surnamestart Le\surnameend}, \bibinfo{author}{Francesco
  \surnamestart Fabiano\surnameend}, \bibinfo{author}{Tran~Cao \surnamestart
  Son\surnameend} \& \bibinfo{author}{Enrico \surnamestart Pontelli\surnameend}
  (\bibinfo{year}{2018}): \emph{\bibinfo{title}{{EFP} and {PG-EFP}: Epistemic
  Forward Search Planners in Multi-Agent Domains}}.
\newblock In: {\sl \bibinfo{booktitle}{Proceedings of the Twenty-Eighth
  International Conference on Automated Planning and Scheduling}},
  \bibinfo{publisher}{{AAAI} Press}, \bibinfo{address}{Delft, The Netherlands},
  pp. \bibinfo{pages}{161--170}.
\newblock
  \urlprefix\url{https://aaai.org/ocs/index.php/ICAPS/ICAPS18/paper/view/17733}.

\bibitemdeclare{inproceedings}{lipovetzky2017best}
\bibitem{lipovetzky2017best}
\bibinfo{author}{Nir \surnamestart Lipovetzky\surnameend} \&
  \bibinfo{author}{Hector \surnamestart Geffner\surnameend}
  (\bibinfo{year}{2017}): \emph{\bibinfo{title}{Best-First Width Search:
  Exploration and Exploitation in Classical Planning}}.
\newblock In: {\sl \bibinfo{booktitle}{Proceedings of the Thirty-First {AAAI}
  Conference on Artificial Intelligence}}, \bibinfo{address}{San Francisco,
  California, {USA}}, pp. \bibinfo{pages}{3590--3596}.
\newblock
  \urlprefix\url{http://aaai.org/ocs/index.php/AAAI/AAAI17/paper/view/14862}.

\bibitemdeclare{inproceedings}{liu2018multi}
\bibitem{liu2018multi}
\bibinfo{author}{Qiang \surnamestart Liu\surnameend} \&
  \bibinfo{author}{Yongmei \surnamestart Liu\surnameend}
  (\bibinfo{year}{2018}): \emph{\bibinfo{title}{Multi-agent Epistemic Planning
  with Common Knowledge.}}
\newblock In: {\sl \bibinfo{booktitle}{{IJCAI} International Joint Conference
  on Artificial Intelligence}}, pp. \bibinfo{pages}{1912--1920}.

\bibitemdeclare{inproceedings}{muise2015planning}
\bibitem{muise2015planning}
\bibinfo{author}{Christian~J. \surnamestart Muise\surnameend},
  \bibinfo{author}{Vaishak \surnamestart Belle\surnameend},
  \bibinfo{author}{Paolo \surnamestart Felli\surnameend},
  \bibinfo{author}{Sheila~A. \surnamestart McIlraith\surnameend},
  \bibinfo{author}{Tim \surnamestart Miller\surnameend},
  \bibinfo{author}{Adrian~R. \surnamestart Pearce\surnameend} \&
  \bibinfo{author}{Liz \surnamestart Sonenberg\surnameend}
  (\bibinfo{year}{2015}): \emph{\bibinfo{title}{Planning Over Multi-Agent
  Epistemic States: {A} Classical Planning Approach}}.
\newblock In: {\sl \bibinfo{booktitle}{Proc. of {AAAI}}}, pp.
  \bibinfo{pages}{3327--3334}.

\bibitemdeclare{book}{Omodeo20171}
\bibitem{Omodeo20171}
\bibinfo{author}{E.G. \surnamestart Omodeo\surnameend},
  \bibinfo{author}{A.~\surnamestart Policriti\surnameend} \&
  \bibinfo{author}{A.I. \surnamestart Tomescu\surnameend}
  (\bibinfo{year}{2017}): \emph{\bibinfo{title}{On sets and graphs:
  Perspectives on logic and combinatorics}}.
\newblock \bibinfo{publisher}{Springer}, \doi{10.1007/978-3-319-54981-1}.

\bibitemdeclare{article}{richter2010lama}
\bibitem{richter2010lama}
\bibinfo{author}{Silvia \surnamestart Richter\surnameend} \&
  \bibinfo{author}{Matthias \surnamestart Westphal\surnameend}
  (\bibinfo{year}{2010}): \emph{\bibinfo{title}{The LAMA planner: Guiding
  cost-based anytime planning with landmarks}}.
\newblock {\sl \bibinfo{journal}{Journal of Artificial Intelligence Research}}
  \bibinfo{volume}{39}, pp. \bibinfo{pages}{127--177}, \doi{10.1613/jair.2972}.

\bibitemdeclare{book}{modernApproach}
\bibitem{modernApproach}
\bibinfo{author}{Stuart \surnamestart Russell\surnameend} \&
  \bibinfo{author}{Peter \surnamestart Norvig\surnameend}
  (\bibinfo{year}{2009}): \emph{\bibinfo{title}{Artificial Intelligence: A
  Modern Approach}}, \bibinfo{edition}{3rd} edition.
\newblock \bibinfo{publisher}{Prentice Hall Press}, \bibinfo{address}{Upper
  Saddle River, NJ, USA}.

\bibitemdeclare{book}{smullyan2012first}
\bibitem{smullyan2012first}
\bibinfo{author}{Raymond~R \surnamestart Smullyan\surnameend}
  (\bibinfo{year}{2012}): \emph{\bibinfo{title}{First-order logic}}.
\newblock \bibinfo{volume}{43}, \bibinfo{publisher}{Springer Science \&
  Business Media}.

\bibitemdeclare{inproceedings}{son2014finitary}
\bibitem{son2014finitary}
\bibinfo{author}{Tran~Cao \surnamestart Son\surnameend},
  \bibinfo{author}{Enrico \surnamestart Pontelli\surnameend},
  \bibinfo{author}{Chitta \surnamestart Baral\surnameend} \&
  \bibinfo{author}{Gregory \surnamestart Gelfond\surnameend}
  (\bibinfo{year}{2014}): \emph{\bibinfo{title}{Finitary S5-theories}}.
\newblock In: {\sl \bibinfo{booktitle}{European Workshop on Logics in
  Artificial Intelligence}}, \bibinfo{organization}{Springer}, pp.
  \bibinfo{pages}{239--252}, \doi{10.1093/jigpal/jzm059}.

\bibitemdeclare{inproceedings}{DBLP:conf/iclp/SonPGB16}
\bibitem{DBLP:conf/iclp/SonPGB16}
\bibinfo{author}{Tran~Cao \surnamestart Son\surnameend},
  \bibinfo{author}{Enrico \surnamestart Pontelli\surnameend},
  \bibinfo{author}{Michael \surnamestart Gelfond\surnameend} \&
  \bibinfo{author}{Marcello \surnamestart Balduccini\surnameend}
  (\bibinfo{year}{2016}): \emph{\bibinfo{title}{An Answer Set Programming
  Framework for Reasoning About Truthfulness of Statements by Agents}}.
\newblock In: {\sl \bibinfo{booktitle}{Technical Communications of the 32nd
  International Conference on Logic Programming, {ICLP} 2016 TCs, October
  16-21, 2016, New York City, {USA}}}, pp. \bibinfo{pages}{8:1--8:4},
  \doi{10.4230/OASIcs.ICLP.2016.8}.

\bibitemdeclare{book}{van2007dynamic}
\bibitem{van2007dynamic}
\bibinfo{author}{Hans \surnamestart Van~Ditmarsch\surnameend},
  \bibinfo{author}{Wiebe \surnamestart van Der~Hoek\surnameend} \&
  \bibinfo{author}{Barteld \surnamestart Kooi\surnameend}
  (\bibinfo{year}{2007}): \emph{\bibinfo{title}{Dynamic epistemic logic}}.
\newblock \bibinfo{volume}{337}, \bibinfo{publisher}{Springer Science \&
  Business Media}.

\bibitemdeclare{inproceedings}{wan2015complete}
\bibitem{wan2015complete}
\bibinfo{author}{Hai \surnamestart Wan\surnameend}, \bibinfo{author}{Rui
  \surnamestart Yang\surnameend}, \bibinfo{author}{Liangda \surnamestart
  Fang\surnameend}, \bibinfo{author}{Yongmei \surnamestart Liu\surnameend} \&
  \bibinfo{author}{Huada \surnamestart Xu\surnameend} (\bibinfo{year}{2015}):
  \emph{\bibinfo{title}{A Complete Epistemic Planner without the Epistemic
  Closed World Assumption}}.
\newblock In: {\sl \bibinfo{booktitle}{{IJCAI} International Joint Conference
  on Artificial Intelligence}}, \bibinfo{address}{Buenos Aires, Argentina}, pp.
  \bibinfo{pages}{3257--3263}.
\newblock \urlprefix\url{http://ijcai.org/Abstract/15/459}.

\end{thebibliography}
%
%
%\newpage
%\appendix
%\input{chLast-appendix}
\end{document}